\documentclass{ELSP}
\PassOptionsToPackage{unicode}{hyperref}
\PassOptionsToPackage{hyphens}{url}
\IfFileExists{upquote.sty}{\usepackage{upquote}}{}
\IfFileExists{microtype.sty}{
	\usepackage[]{microtype}
	\UseMicrotypeSet[protrusion]{basicmath} 
}{}
\makeatletter
\IfFileExists{parskip.sty}{%
	\usepackage{parskip}
}
\makeatother
\usepackage{xcolor}
\IfFileExists{xurl.sty}{\usepackage{xurl}}{} 
\IfFileExists{bookmark.sty}{\usepackage{bookmark}}{\usepackage{hyperref}}
\hypersetup{
	pdfcreator={ELSP}}
\usepackage{longtable,booktabs,array}
\usepackage{calc} 
\usepackage{etoolbox}
\makeatletter
\patchcmd\longtable{\par}{\if@noskipsec\mbox{}\fi\par}{}{}
\makeatother
\IfFileExists{footnotehyper.sty}{\usepackage{footnotehyper}}{\usepackage{footnote}}
\makesavenoteenv{longtable}
\usepackage{setspace}
\usepackage{graphicx}
\usepackage{amsmath}
\usepackage{times}
\usepackage{mathptmx}
\usepackage{color}
\usepackage{caption}
\usepackage{subcaption}
\usepackage{geometry}
\usepackage{float}
\usepackage{enumerate}
\usepackage{stfloats}
\usepackage{ragged2e}

\usepackage[markup=underlined]{changes}
\usepackage{multirow}

\usepackage{fancyhdr}
\fancypagestyle{firstpage}
{
	\fancyhf{}
	\setlength{\headsep}{6px}
	\fancyhead[R]{\sffamily \fontsize{9pt}{8pt}\selectfont \textbf{\textcolor[RGB]{0,131,255}{Robot Learn.}}}
	\fancyhead[L]{\sffamily \fontsize{9pt}{8pt}\selectfont \textbf{\textcolor[RGB]{0,131,255}{ELSP}}}
	\fancyfoot[L]{\sffamily \fontsize{9pt}{8pt}\selectfont DeFazio D, {\em et al}. Robot Learn. 2025(Issue):XXXX}
	
}

\hypersetup{
    colorlinks=false, %
    pdfborder={0 0 0}, %
    hidelinks %
}

\usepackage{cite}
\usepackage[none]{hyphenat}

\usepackage[utf8]{inputenc}
\usepackage[table,xcdraw,RGB]{xcolor}
\usepackage{enumitem}
\usepackage{soul,xcolor} 
\sethlcolor{yellow} 

\usepackage[numbers,sort,compress]{natbib}
\usepackage{natbib}
\setcitestyle{numbers,square,comma,sort&compress}
\newlength\mylen
\setcitestyle{citesep={,\kern-\mylen}}
\begin{document}

\thispagestyle{firstpage}

\let\thefootnote\relax
\footnotetext{
\newline
\newline
	\begin{minipage}[h]{0.15\linewidth}
	\includegraphics{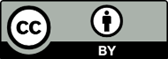}
	\end{minipage}
	\hfill
	\begin{minipage}[h]{0.8\linewidth}
		\footnotesize{Copyright©2025 by the authors. Published by ELSP. 
			This work is licensed under a Creative Commons Attribution 4.0 
			International License, which permits unrestricted use, distribution, 
			and reproduction in any medium provided the original work is properly cited.}
	\end{minipage}

}

\begin{spacing}{0.88}
{\sffamily\fontsize{10.5pt}{10.5pt}\selectfont
\raggedright
\noindent{\textls[-20]{Article $\mid$ Received Day Mon Year; Revised Day Mon Year; Accepted 6 November 2025; Published Day Mon Year}}}\\
{\sffamily\small{https://doi.org/10.55092/rl2025xxxx}}
\end{spacing}

\vspace{+18pt}
\papertitle{Vision language models can parse floor plan maps}
\vspace{+12pt}


\vspace{-20pt}
\hspace{-0.92cm}\authorname{David} {DeFazio} {}\textbf{†,}
\authorname{Hrudayangam} {Mehta} {}\textbf{†,}
\authorname{Meng} {Wang} {}\textbf{†,}
\authorname{Ping} {Yang}{}\textbf{,}
\authorname{Jeremy} {Blackburn} {}
\textbf{and} \\
\authornameCorres{Shiqi} {Zhang}{}\textbf{*}

\vspace{+8pt}
\hspace{-1.08cm}\formatintroduction{}{School of Computing, Binghamton University, Binghamton, New York, USA}

\vspace{-16pt}
\noindent \hangindent=0.8em 
\hangafter=1 \textnormal{\textdagger} \hspace{+0.1cm}These authors contributed equally to this work.

\vspace{+10pt}
\hspace{-0.8cm}\authoremail{Correspondence author} {zhangs@binghamton.edu.}\\

\vspace{-8pt}
\noindent\textbf{\textcolor[RGB]{0,131,255}{Highlights:}}
\begin{itemize}[left=0pt, labelwidth=0pt, labelsep=17pt, itemsep=0pt]
\vspace{+6pt}
    \item This article shows that current vision-language models (VLMs) can be used for parsing floor plan maps and planning for indoor navigation tasks. 
    \item The proposed system was evaluated using real floor plan maps and two multimodal foundation models, including GPT-4o and Claude-3.5 Sonnet. 
    \item Results demonstrated the effectiveness of VLM-based robot navigation systems, highlighted practical suggestions, and illustrated applications on a quadruped robot. 
\end{itemize}

\vspace{+12pt}
\noindent\textbf{\textbf{\textcolor[RGB]{0,131,255}{Abstract:}}} 
Vision language models (VLMs) can simultaneously reason about images and texts to tackle many tasks, from visual question answering to image captioning. 
This paper focuses on map parsing, a novel task that is unexplored within the VLM context and particularly useful to mobile robots. 
Map parsing requires understanding not only the labels but also the geometric configurations of a map, \textit{i.e.}, what areas are like and how they are connected. 
To evaluate the performance of VLMs on map parsing, we prompt VLMs with floor plan maps to generate task plans for complex indoor navigation. 
Our results demonstrate the remarkable capability of VLMs in map parsing, with a success rate of 0.96 in tasks requiring a sequence of nine navigation actions, e.g., approaching and going through doors.
Other than intuitive observations, e.g., VLMs do better in smaller maps and simpler navigation tasks, there was a very interesting observation that its performance drops in large open areas. 
We provide practical suggestions to address such challenges as validated by our experimental results.
Webpage: \url{https://sites.google.com/view/vlm-floorplan/}.

\vspace{+12pt}
\noindent\textbf{\textcolor[RGB]{0,131,255}{Keywords:}} task planning; mobile robots; vision language models; robot navigation

\section{Introduction}
\label{sec:intro}

\textls[-2]{A key to mobile robotics is a deep understanding of the geometric configuration of the world that mobile robots live in. 
As a result, many mobile robots need some forms of a map 
for localization, obstacle avoidance and navigation. 
To build such maps, the robots use a predefined data structure, e.g., an occupancy grid~\cite{elfes1989using} or visual features~\cite{taketomi2017visual}, and then perform simultaneous localization and mapping~(SLAM). 
Human beings have a long history of building and using maps. 
These days one can easily read floor plan maps of airports and shopping centers, and figure out a plan for navigation. 
By comparison, robots can hardly reach comparable competency in map reading and task planning. 
It is a non-trivial task for the robots due to the many labels in the map, their ambiguous associations to different areas, and complex geometric configurations. 
As a result, there is no existing method for addressing the map parsing problem, \textit{i.e.}, computing a navigation plan given a map image and a goal text, which motivated this research.} 

For complex navigation tasks, a robot needs to compute a task plan, \textit{i.e.}, a sequence of navigation actions, and continuous  trajectories for realizing those actions. 
Example actions can be entering an area and going through a door. 
Extensive engineering efforts are needed to realize such navigation systems, from building the map itself to labeling areas of the map. 
At the same time, professional architectural drafters have generated blueprints that accurately reflect the geometric configurations, which unfortunately cannot be used by current robots. 
From an application perspective, this research aims to leverage the readily accessible floor plan maps in human environments to fulfill the robots' need of maps for navigation. 

\begin{figure}[t]
    \centering
    \vspace{.5em}
    \includegraphics[width=0.66\textwidth]{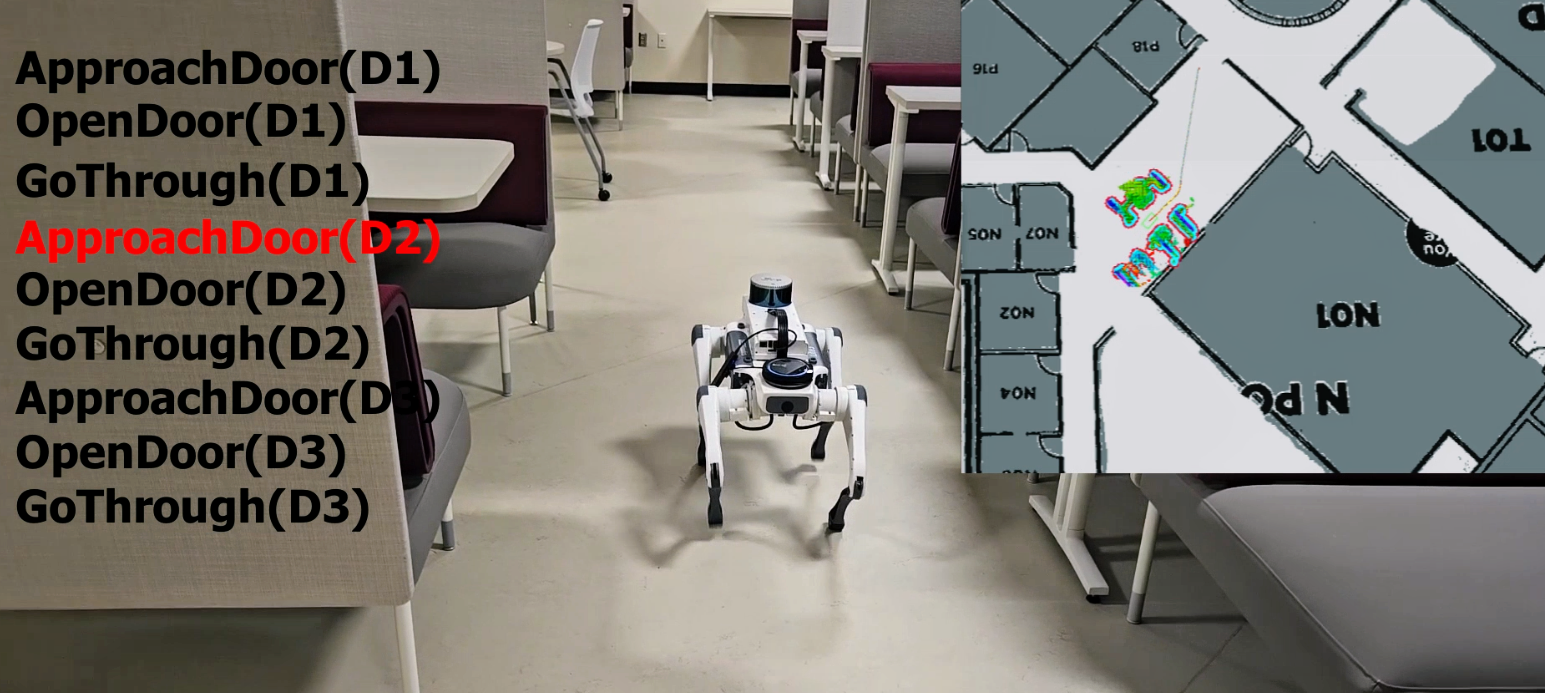}
    \caption{\textls[-12]{Quadruped robot executing a VLM-generated plan (current action highlighted in red) to complete a navigation task while localizing directly on a floor plan image.}}
    \vspace{-.3em}
    \label{fig:demo_pic}
\end{figure}

Vision language models (VLMs) are foundation models that learn from and reason about both images and texts, supporting a variety of downstream tasks from visual question answering to image captioning. 
VLMs have demonstrated impressive successes in a variety of applications~\cite{antol2015vqa, chen2020uniter, brooks2023instructpix2pix}. 
Among them, robotics is an important application domain~\cite{nasiriany2024pivot, liu2024moka,hu2023toward}. 
We, for the first time, apply VLMs to the novel task of map parsing and evaluate its performance in navigation, a foundational problem in mobile robotics. 
Our approach is simple and intuitive. 
A floor plan map and a problem description, including the start and goal positions, are provided to a VLM, and the task is to compute a plan (\textit{i.e.}, an action sequence) to achieve the goal. 
Figure~\ref{fig:demo_pic} shows a quadruped mobile robot executing a navigation plan that was generated by a VLM using a floor plan map. 
Despite the straightforward idea, the results are surprisingly good.
Navigation plans generated by VLMs which can require a sequence of nine actions are generated correctly up to 90\% of the time on some floor plans.

The main contribution of this research includes the introduction of the map parsing problem, evaluations of VLMs on this problem, practical suggestions that paves the way for future research on VLM-based map parsing, and a complete demonstration of real-robot system. 

\vspace{-4pt}
There are limitations in this research that can be addressed in future work. 
One is that we still need to slightly edit the map image, such as thickening walls and removing architectural annotations, to produce the best performance. 
Such steps can be automated in future work. 
Another is that the robot needs to stand close and forward-facing when capturing the map image. 
This can be a challenge to small robots because most floor plan maps are placed at human heights. 
In this paper, we focus on highlighting the remarkable performance of VLMs on map parsing, and leave those topics to future work.

\section{Related work}

In this section, we discuss existing work in autonomous navigation for mobile robots, VLM prompting strategies, and integrating large pre-trained models in robotics. We highlight how our work differs from existing works in each of these
categories.

\vspace{-6pt}
\subsection{Existing map representations and navigation}

Existing works demonstrating autonomous navigation on mobile robots can be categorized into two groups. 
One relies on generating an occupancy grid map~\cite{zhang2015mobile,hayamizu2021guiding,fu2022coupling,defazio2023seeing}, where each pixel in the map identifies an area being occupied, obstacle-free or unknown. 
Another popular way of realizing mobile navigation leverages vision-based methods for simultaneous localization and mapping (Visual SLAM)~\cite{taketomi2017visual, zhang2021frl, macario2022comprehensive, sorokin2022learning, hwang2023system}. 
Those methods rely on visual features, and then apply simultaneously estimate the locations of both visual landmarks and the robot itself~\cite{thrun2002probabilistic}. 
While such map representations have proven to be effective for autonomous navigation tasks, generating an accurate map is oftentimes labor-intensive. 
For instance, in vision-based settings, the robot has limited knowledge of the global environment, either leading to lengthy exploration, or navigational commands from a human. 
In this work, we greatly reduce the effort of generating accurate maps while still leveraging detailed information of the environment through the use of existing, and potentially \textit{in situ}, floor plans.

\vspace{-4pt}
\subsection{VLM prompting strategies}

The output of a large pre-trained model largely relies on the way it is prompted. Strategies like chain-of-thought~\cite{wei2022chain}, in-context learning~\cite{dong2022survey} and others~\cite{yao2023tree,besta2024graph} are leveraged on LLMs to improve performance. \textls[+20] {Similar to language prompts, image prompting strategies can improve VLM outputs. Set-of Mark prompting, which segments and labels objects in an image has shown to improve VLM responses~\cite{yang2023set}. In our work, we design a new visual prompting strategy designed for obtaining a spatial understanding of floor plan images.}

\vspace{-4pt}
\subsection{Large models in robotics}

To improve the common-sense reasoning capabilities of robots, large pre-trained LLMs and VLMs have been integrated into robotic systems for various tasks. 
Housekeeping~\cite{kant2022housekeep,li2023behavior} and object rearrangement~\cite{ding2023task,huang2019large,danielczuk2021object} are two of the service task examples. Navigation is another robotics domain where LLMs and VLMs have demonstrated exceptional performance~\cite{yokoyama2024vlfm, rajvanshi2024saynav, song2024socially, sathyamoorthy2024convoi, chen2024commonsense}. 
Furthermore, while quadruped locomotion problems traditionally were solved using optimization methods, the effectiveness of LLMs and VLMs has been evident in recent years~\cite{yu2023language, tang2023saytap,gu2025humanoid}.
Alignment of the large model with the environment and robot's skills is critical to perform tasks in the real world~\cite{ahn2022can}. \mbox{Various approaches} have demonstrated planning capabilities~\cite{huang2022inner, liu2023llm+} and uncertainty estimation~\cite{ren2023robots} of large models. Various visual prompting strategies designed for different robotic manipulation tasks have also been \mbox{developed~\cite{nasiriany2024pivot, liu2024moka}}. In line with recent works that leverage large models to incorporate common-sense knowledge on robots, we generate feasible navigation plans directly from an image of a floor plan.

\section{Methodology}
\label{sec:methodology}

In this section, we present our approach for leveraging Vision-Language Models (VLMs) to interpret floor plan images and generate navigation instructions. 
We discuss the two key aspects of our approach: visual prompting strategy, and VLM-based plan generation.

\vspace{-4pt}
\subsection{Visual prompting strategy}
\label{sec:prompt}

Our study uses floor plan images that include detailed architectural layouts for various building types.
To generate accurate plans from a VLM, it's important to design a visual prompt which can facilitate learning the structure of the floor plan.
However, raw floor plan images tend to contain various markings (\textit{i.e.} windows, furniture symbols, non-uniform wall thickness) which can potentially confuse the VLM in understanding the general layout of the floor plan.
Thus, we remove such markings to produce a cleaner map which can be better leveraged by a VLM.

We find that removing extraneous details from the floor plan is insufficient for the VLM to understand the map layout. In particular, we find the VLM has limited spatial awareness for sparsely labelled rooms with lots of open space, and near key decision points like doors and intersections. To alleviate this, we add duplicate room labels in open spaces and near doors and intersections. This provides the VLM with further guidance in understanding the general structure of the floor plan. We later demonstrate the importance of such additional labels in Section~\ref{sec:experiments} 

In this study, all floor plans are preprocessed through a simple yet systematic enhancement procedure, Next, we present the step-by-step instructions that we followed for map enhancement. 

\begin{enumerate}[label=(\arabic*)]\setlength{\leftskip}{+0.48cm}
    \item Visual simplification: Raw floor plans frequently include architectural annotations outside the map itself, such as dimensions and scale bars. We remove those visuals and keep only the map part of the floor plans. In addition, there are non structural visuals, such as window icons, and furniture, where people use very different icons or symbols. 
    Those were removed for \mbox{standardizing evaluations}. 

    \item Structural standardization: The line segments for walls and doors can be different in floor plan maps. 
    We adjust the wall thickness to a uniform value. 
    Doors are colored yellow, making their locations visually salient and uniformly represented across maps. 
    Windows are replaced with walls to avoid confusing the robots on whether windows are navigable. 

    \item Dense labeling: Some rooms are very large, of irregular shape (e.g., long corridors), or both. 
    We added additional labels of rooms at areas near doors or intersections, or within large open areas. 
    This dense labeling provides additional contextual cues, leading to significant performance improvements, especially in complex environments involving multiple rooms and transitions. 

    \item \textls[-10]{Consistency check: The last step is to ensure that all rooms are enclosed, no overlapping or misleading elements remain, and every object (rooms and doors in particular) has a unique ID. 
    The finalized map is then exported as a high-resolution PNG image as the image prompt of \mbox{the VLMs}}.
\end{enumerate}

In this work, the map enhancement steps were manually performed, though the process is straightforward and can be reproduced by robotics practitioners using our instructions. 
We discuss possible ways of automating the map enhancement process near the end of this article as future work.

\subsection{VLM-based plan generation}

\begin{figure}[t]
    \centering
    \vspace{1em}
    \hspace{1.5em}
    \includegraphics[width=.97\textwidth]{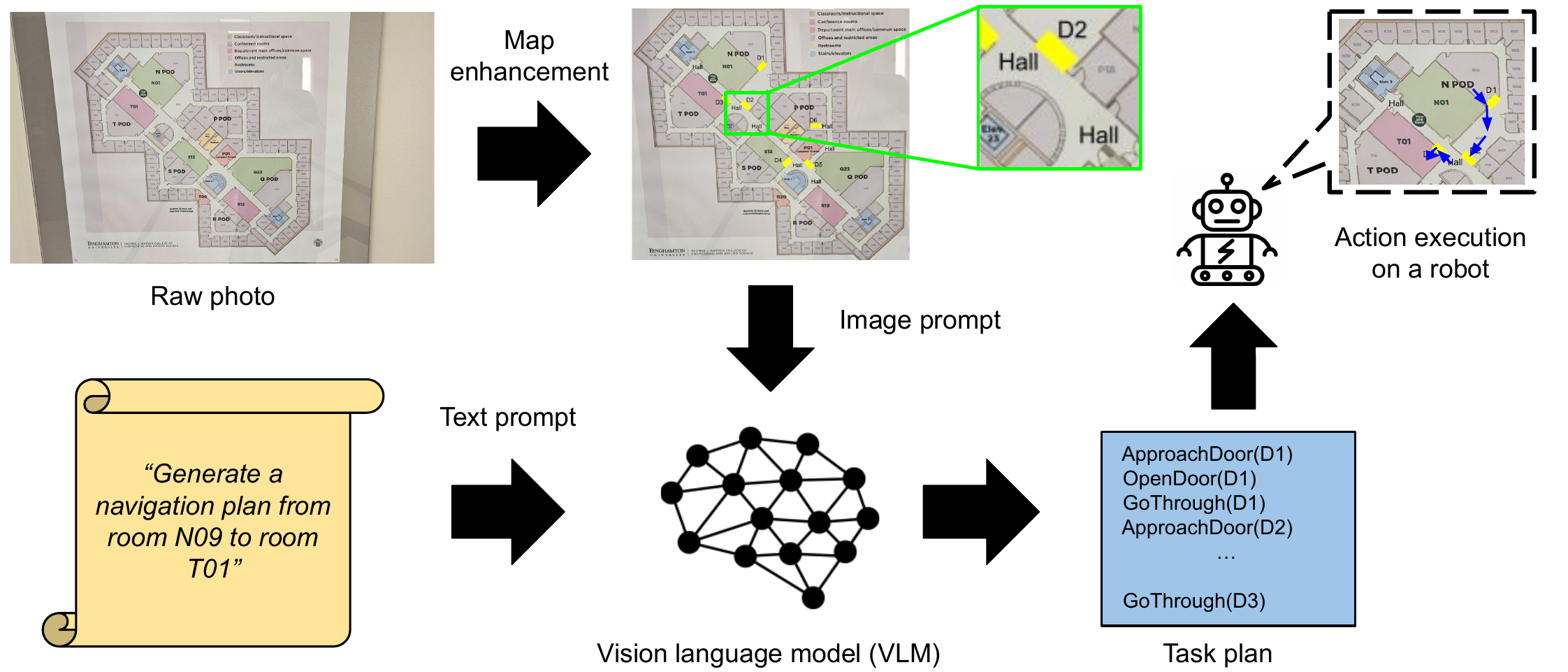}
    \caption{Overview of our method. A robot takes a raw image of a floor plan, which is then enhanced with labels and door indicators. The enhanced floor plan, along with a text prompt specifying the start and goal locations is given to a VLM. The VLM generates a navigation plan to reach the goal location, and the plan is executed on a mobile robot.}
    \label{fig:overview}
\end{figure}

In our VLM-based framework, as shown in Figure~\ref{fig:overview}, VLMs formulate navigation plans based on a floor plan image and a text prompt. This method leverages an instruction-based text prompting strategy, which involves providing the VLM with explicit and detailed guidelines for the navigation task, along with the floor plan image. We now elaborate on the prompting strategy and the resulting output format.

\begin{figure}[H]
    \centering
\includegraphics[width=0.78\textwidth]{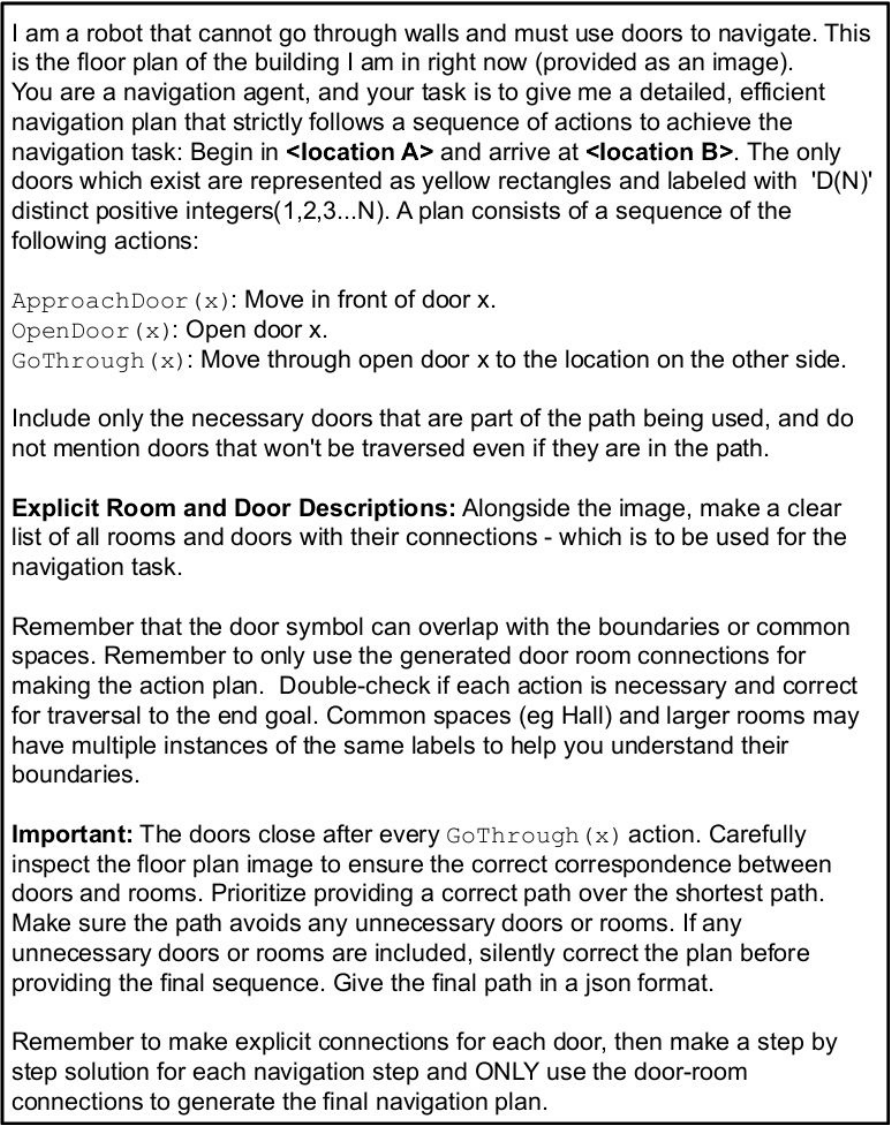}
    \caption{Text prompt input to VLM to generate navigation plans. We \mbox{define the} starting and ending locations, action types, and ask for explicit room and door connections to gain insights as to how the VLM understands the map. }
    \vspace{-.5em}
    \label{fig:prompt_fig}
\end{figure}

The text prompt given to the VLM is shown in Figure~\ref{fig:prompt_fig}. The prompt explicitly defines the starting point and the destination. This allows the VLM to understand the required navigation path and objectives clearly. It provides detailed instructions on interpreting the floor plan and managing door interactions. These guidelines include specific protocols for door operations and decision-making processes.

\vspace{-4pt}

A key aspect of this prompt is the request for all door and room connections. By generating this information at the start, we speculate that the VLM integrates it with the floor plan image to produce an accurate navigation sequence. \textls[-10]{We believe this step helps the VLM better understand the spatial relationships in the map, leading to accurate navigation path planning.}

Based on the text prompt and floor plan image, the VLM generates a sequence of actions required to navigate from the initial to goal location. This sequence includes specific steps, e.g., approaching, opening, and passing through doors.

The output from the VLM is a detailed sequence of actions formatted as follows:
\begin{itemize}\setlength{\leftskip}{+0.16cm}
    \item \texttt{ApproachDoor(x):} Move in front of door \( x \).
    \item \texttt{OpenDoor(x):} Open door \( x \).
    \item \texttt{GoThrough(x):} Move through open door \( x \) to the location on the other side.
\end{itemize}

The final navigation plan is output in JSON format, specifying each action. This structured format facilitates easy interpretation and execution of the navigation instructions. This plan is then parsed and executed by the robot.

\section{Experiments}
\label{sec:experiments}

In order to evaluate whether the VLM produces accurate navigation plans, we design and run experiments over a dataset of floor plans. The experiments are designed to measure the effect of floor plan size, task difficulty, and label density on the VLM's plan accuracy.

\subsection{Experimental setup}

\begin{figure}[t]
    \centering
    \includegraphics[width=0.95\linewidth]{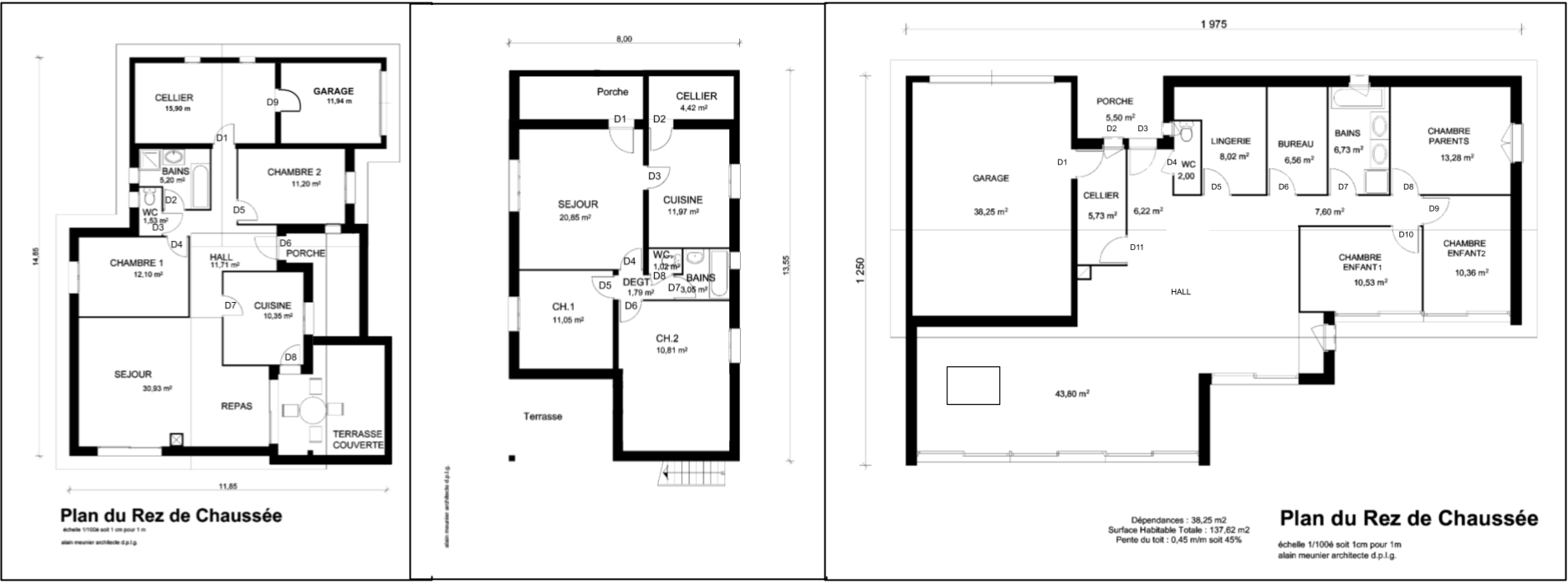}
    \caption{Three raw floor plan maps.}
    \label{fig:raw-maps}
\end{figure}

\begin{figure}[t]
    \centering
\includegraphics[width=0.99\textwidth]{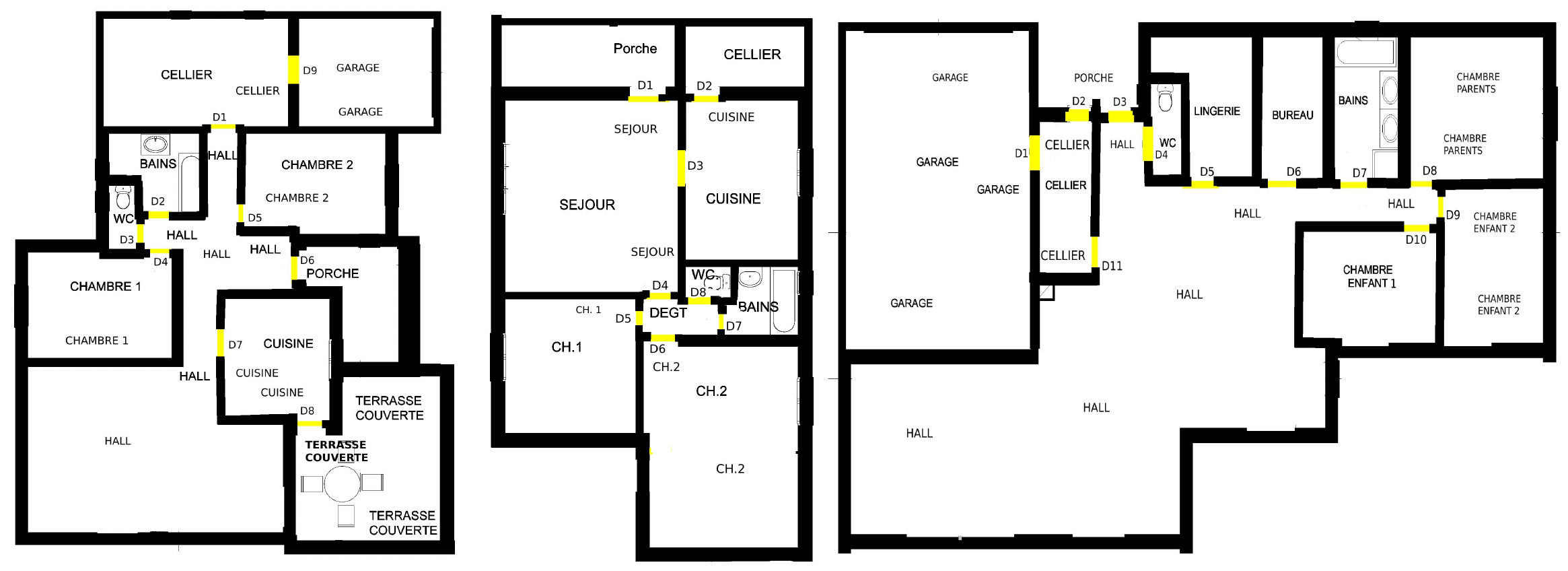}
    \caption{Three maps used in our experiments for evaluating the performance of VLMs in map parsing and plan generation.}
    \label{fig:all_maps}
\end{figure}

Our study uses floor plan images from a publicly available dataset CVC-FP~\cite{Heras15a}, 
which includes detailed architectural floor plans for various building types. 
Three floor plans were randomly selected from those that contain 9--11 rooms and clear labels.
Those raw maps are shown in Figure~\ref{fig:raw-maps}. As illustrated, these images often contain irrelevant visual elements such as scale bars, dimension lines, and furniture/window symbols, which are not pertinent to the general layout. To facilitate more effective visual prompting for VLMs, we preprocess these raw maps using the procedure described in Section~\ref{sec:prompt}
The resulting maps are referred to as ``original maps” and are shown in Figure~\ref{fig:all_maps}. Our experiments use two state-of-the-art VLMs: GPT-4o~\cite{GPT-4o} and Claude-3.5 Sonnet~\cite{Claude35}. 

For each floor plan, we generate five pairs of start and goal locations. To account for the stochastic nature of VLM responses, we run each VLM ten times per navigation task, resulting in a total of 50 navigation trials per floor plan. 
We evaluate performance based on the correctness of the generated plans. 
A plan is considered correct if it uses only those actions defined in the text prompt, includes feasible actions, and leads a sequence of transitions from the start location to the goal. 
Example infeasible actions include navigating to a room that is not connected to the current room and opening a door that belongs to a distant room.

Our experimental design focuses on three key dimensions: 


\begin{enumerate}[label=(\arabic*)] \setlength{\leftskip}{+0.48cm}
    \item Map size: We hypothesize that increasing the map size will result in a decrease in VLM's map parsing performance. This hypothesis is based on the assumption that larger maps introduce greater complexity. 
    
    \item Task difficulty: \textls[-20]{We hypothesize that when the start and end locations are far away (based on number of rooms required to traverse), the VLMs will have a low accuracy in map parsing and \mbox{plan generation}. }

    
    \item Label density: We hypothesize that a densely labelled floor plan map will facilitate accurate navigation plan generation from VLMs. 
    
\end{enumerate}

Next, we describe our experiment setup for evaluating each of the three hypotheses. 

\subsubsection{Map size}

\begin{figure}[t]
    \centering
    \vspace{.8em}
    \includegraphics[width=0.68\textwidth]{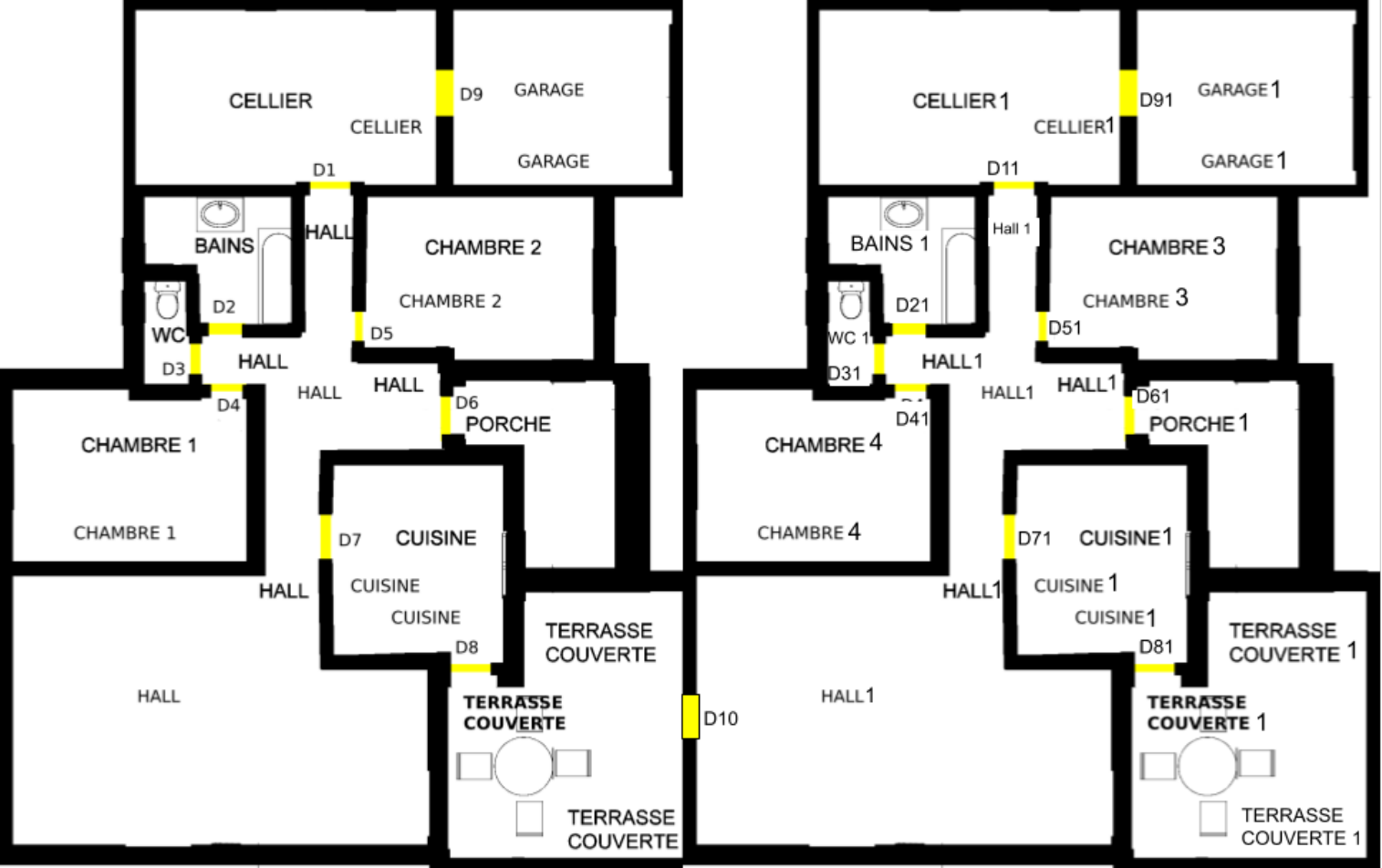}
    \caption{Example of a doubled map. }
    \label{fig:doubled_map}
\end{figure}

To examine the impact of map size on navigation performance, we developed two types of maps: Original Maps that are enhanced versions of the maps selected from the CVC-FP dataset, and Doubled Maps that were created by connecting two copies of an original map through an additional door. 
Figure~\ref{fig:doubled_map} presents an example of a doubled map. 
A door denoted as D10 is added to establish connectivity, thereby forming the final doubled map.


\subsubsection{Task difficulty}

We design two types of navigation tasks to evaluate model performance across two levels of difficulty. 
Easy Tasks are the navigation tasks that can be completed by navigating from Room A to Room B without traversing any intermediate rooms. 
Hard Tasks are those that require a robot to navigate through at least two intermediate rooms before the goal can be achieved. 
As a result, an optimal solution of hard tasks will involve four rooms in total. 
The increased complexity introduces more decision points and possible paths, producing a more challenging task.
    
\vspace{-4pt}
\subsubsection{Label density}

We evaluated the impact of label density on model performance by implementing two labeling schemes. 
Sparse-Labeled maps are those where each room was labeled with a single label, usually placed at the center. 
This minimalistic approach offered fewer cues for the model to base its navigation decisions on. 
Dense-Labeled maps include multiple labels for each room, where the placement is described in Section~\ref{sec:prompt}

\subsection{Experimental results}

\vspace{+2pt}
\subsubsection{Hypothesis 1 (map size)}

The first hypothesis explores how the size of the map influences the model's accuracy. We compared the performance between original maps and doubled maps over hard tasks to assess the same.

\begin{figure}[t]
    \centering
    \begin{subfigure}[b]{0.32\textwidth}  
        \centering
        \includegraphics[width=\textwidth]{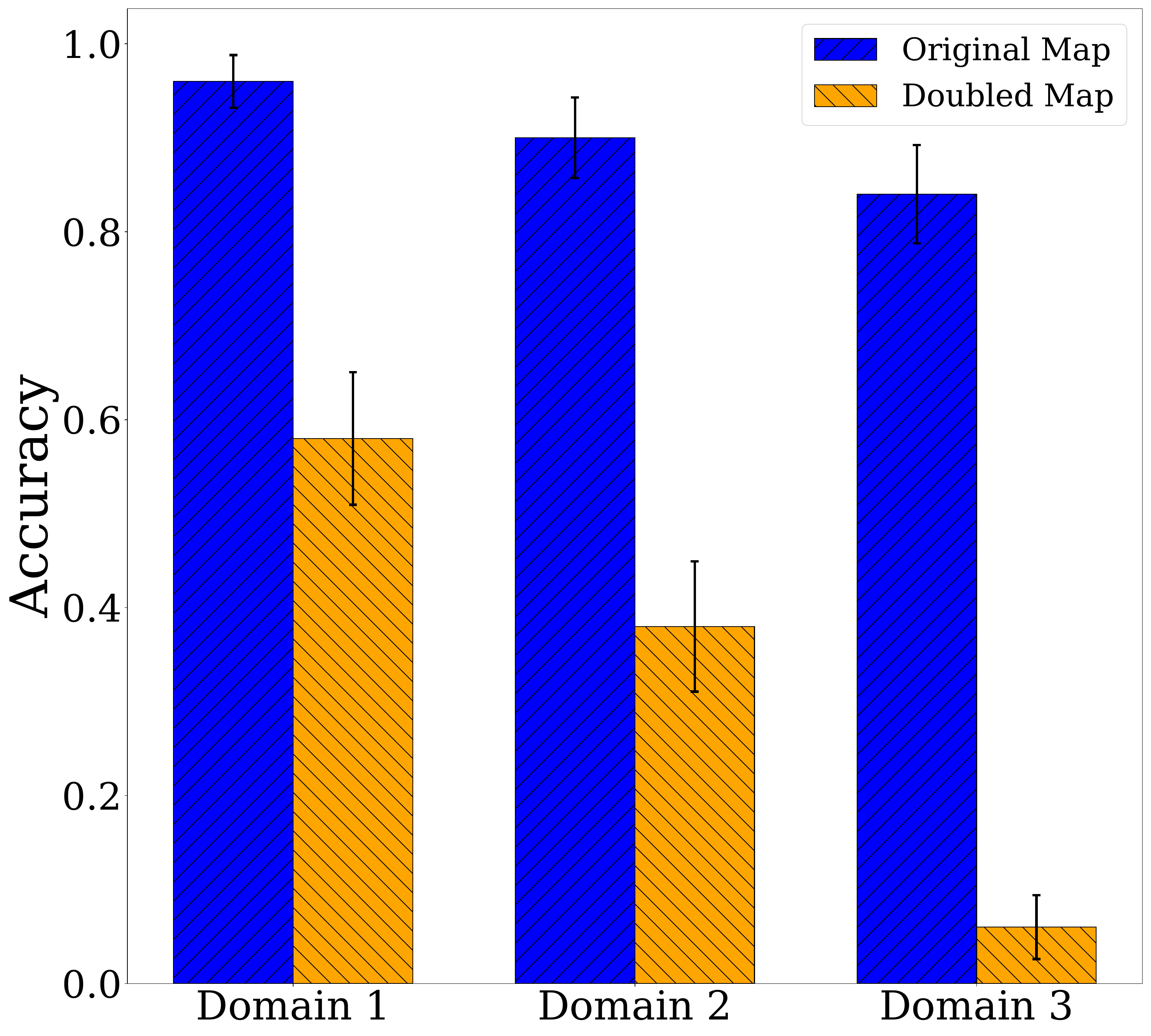}
        \caption{}
        \label{fig:map_size}
    \end{subfigure}
    \begin{subfigure}[b]{0.32\textwidth}  
        \centering
        \includegraphics[width=\textwidth]{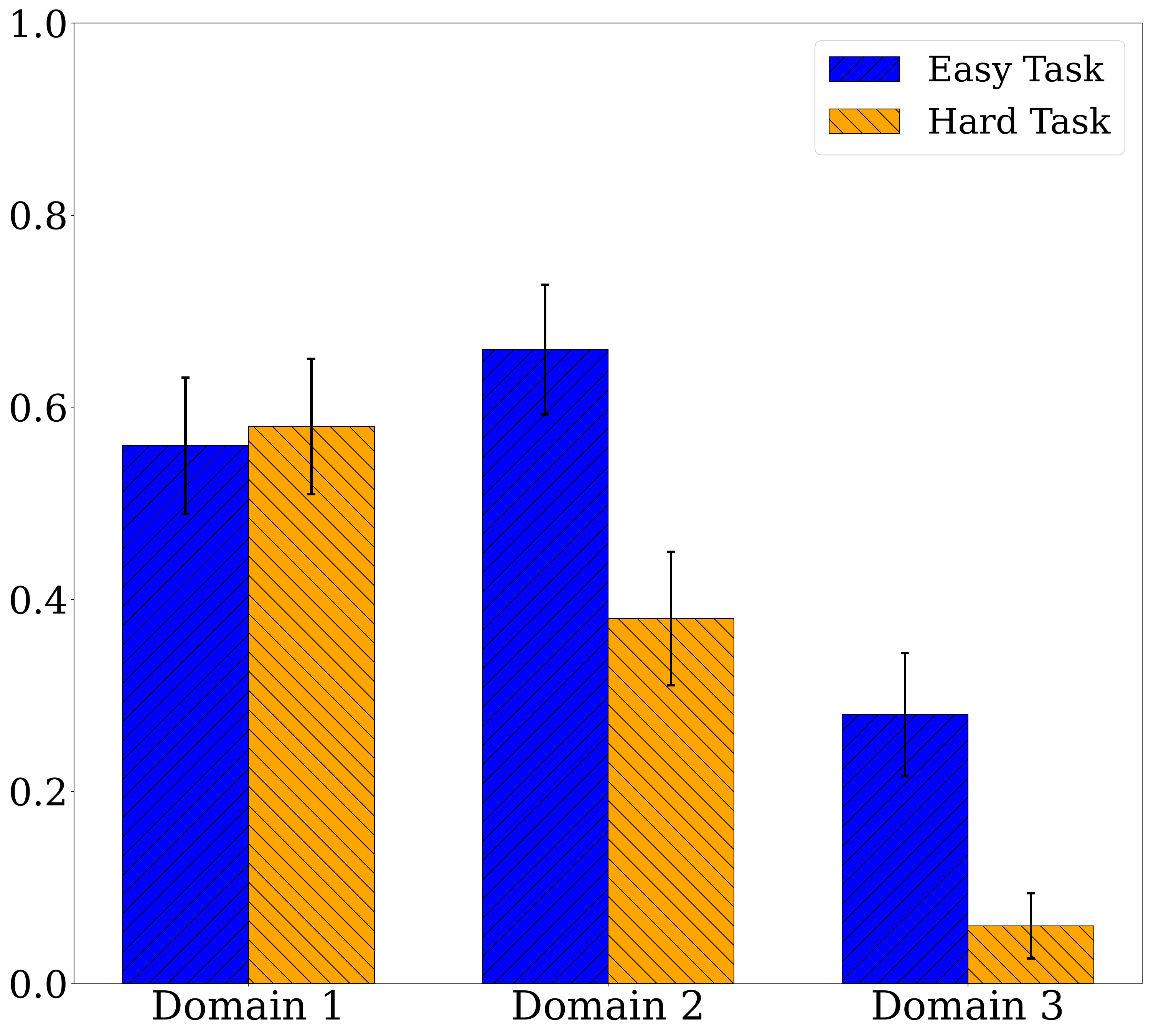}
        \caption{}
        \label{fig:task_dif}
    \end{subfigure}
    \begin{subfigure}[b]{0.32\textwidth}  
        \centering
        \includegraphics[width=\textwidth]{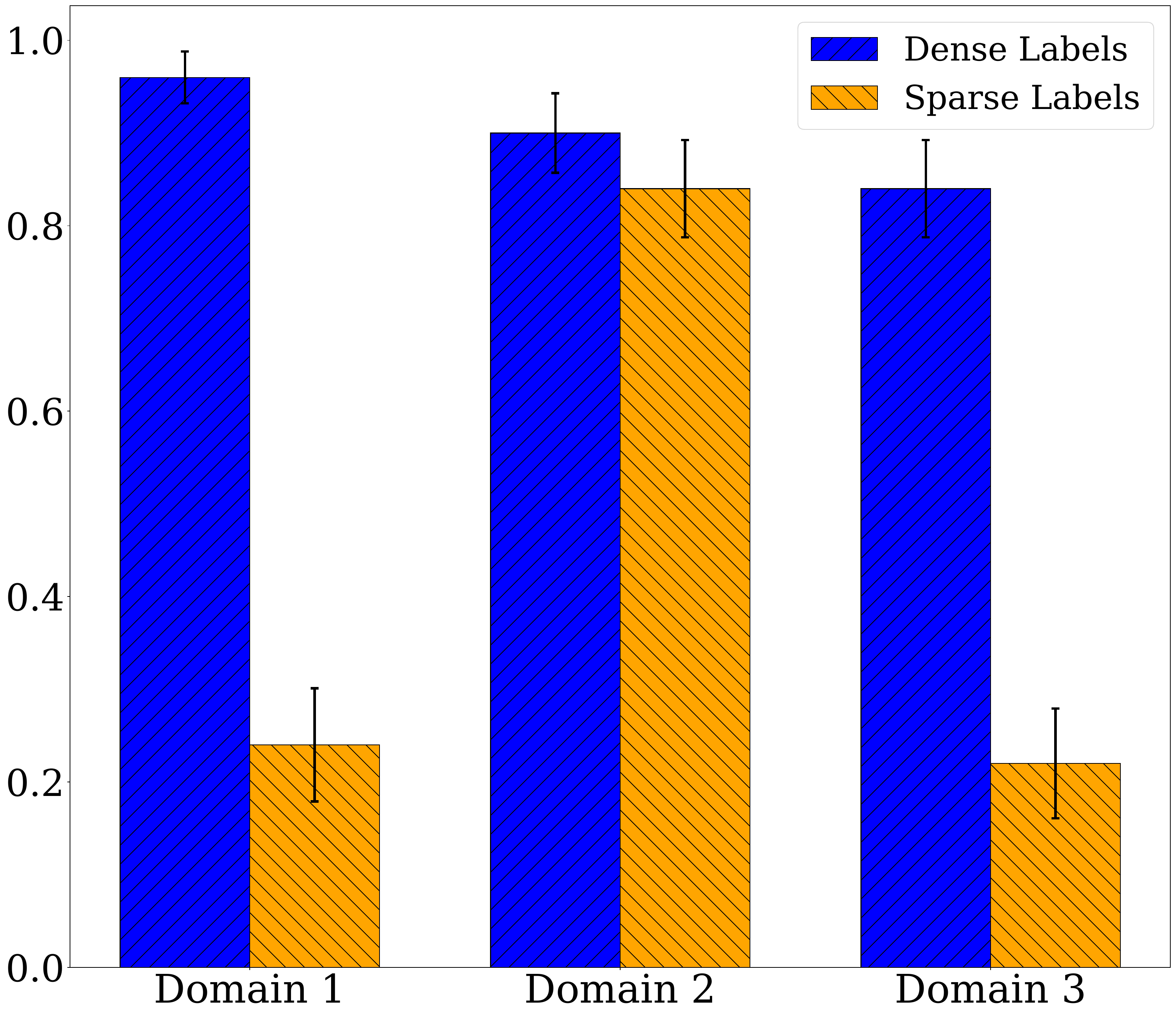}
        \caption{}
        \label{fig:lab_den}
    \end{subfigure}

    \caption{Comparison of GPT-4o results across three experimental conditions: \mbox{\textbf{(a)} Map size:} 
    performance on original maps \textit{vs.} doubled maps over hard tasks; \textbf{(b)} Task difficulty: performance on doubled maps over easy \textit{vs.} hard tasks; \textbf{(c)} Label density: performance on original maps annotated with sparse \textit{vs.} dense labels over hard tasks.} 
    \label{fig:gpt_comparison_horizontal}
\end{figure}

\begin{figure}[t]
    \centering
    
    \begin{subfigure}[b]{0.32\textwidth}  
        \centering
        \includegraphics[width=\textwidth]{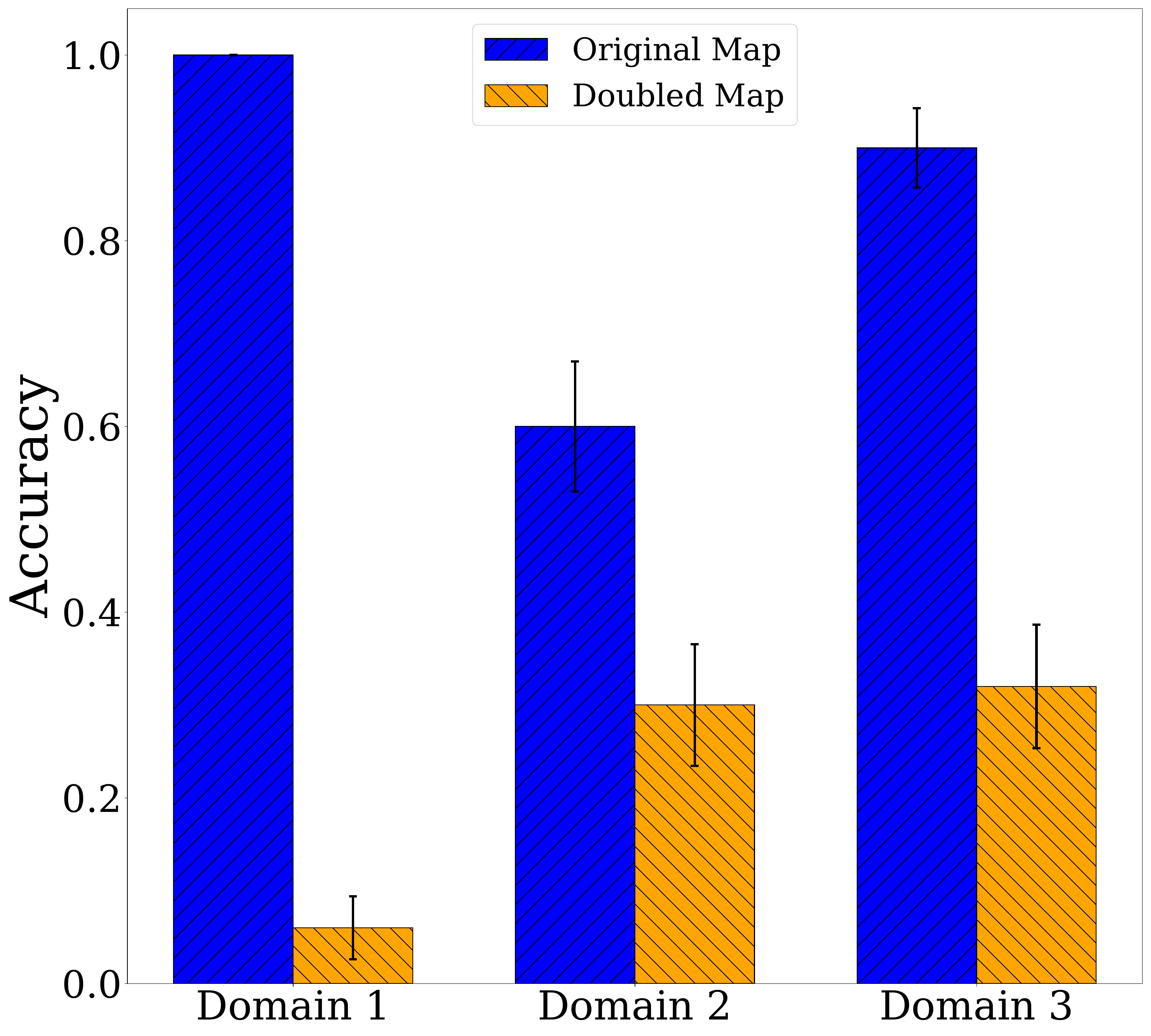}
        \caption{}
        \label{fig:map_size}
    \end{subfigure}
    \begin{subfigure}[b]{0.32\textwidth}  
        \centering
        \includegraphics[width=\textwidth]{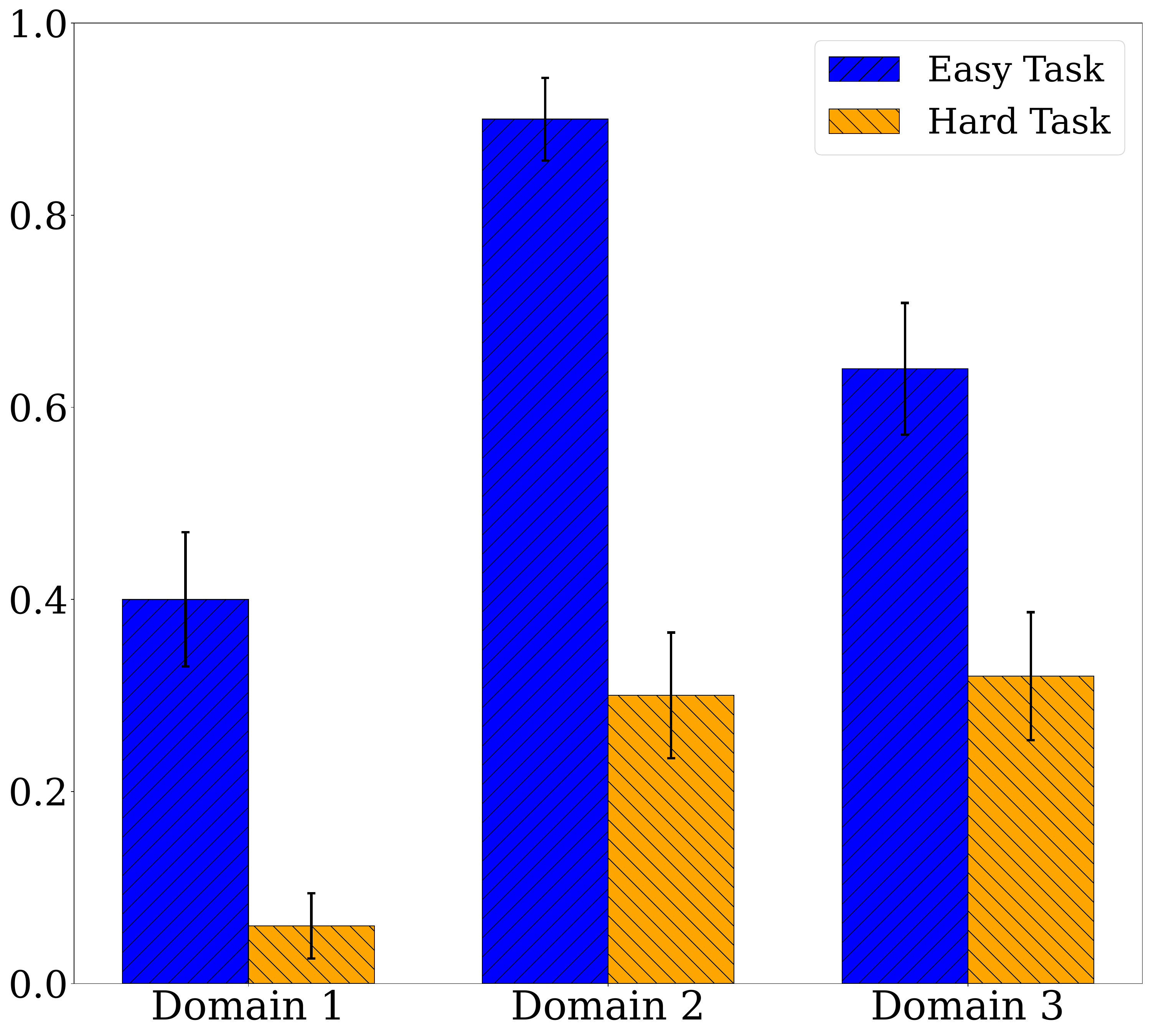}
        \caption{}
        \label{fig:task_dif}
    \end{subfigure}
    \begin{subfigure}[b]{0.32\textwidth}  
        \centering
        \includegraphics[width=\textwidth]{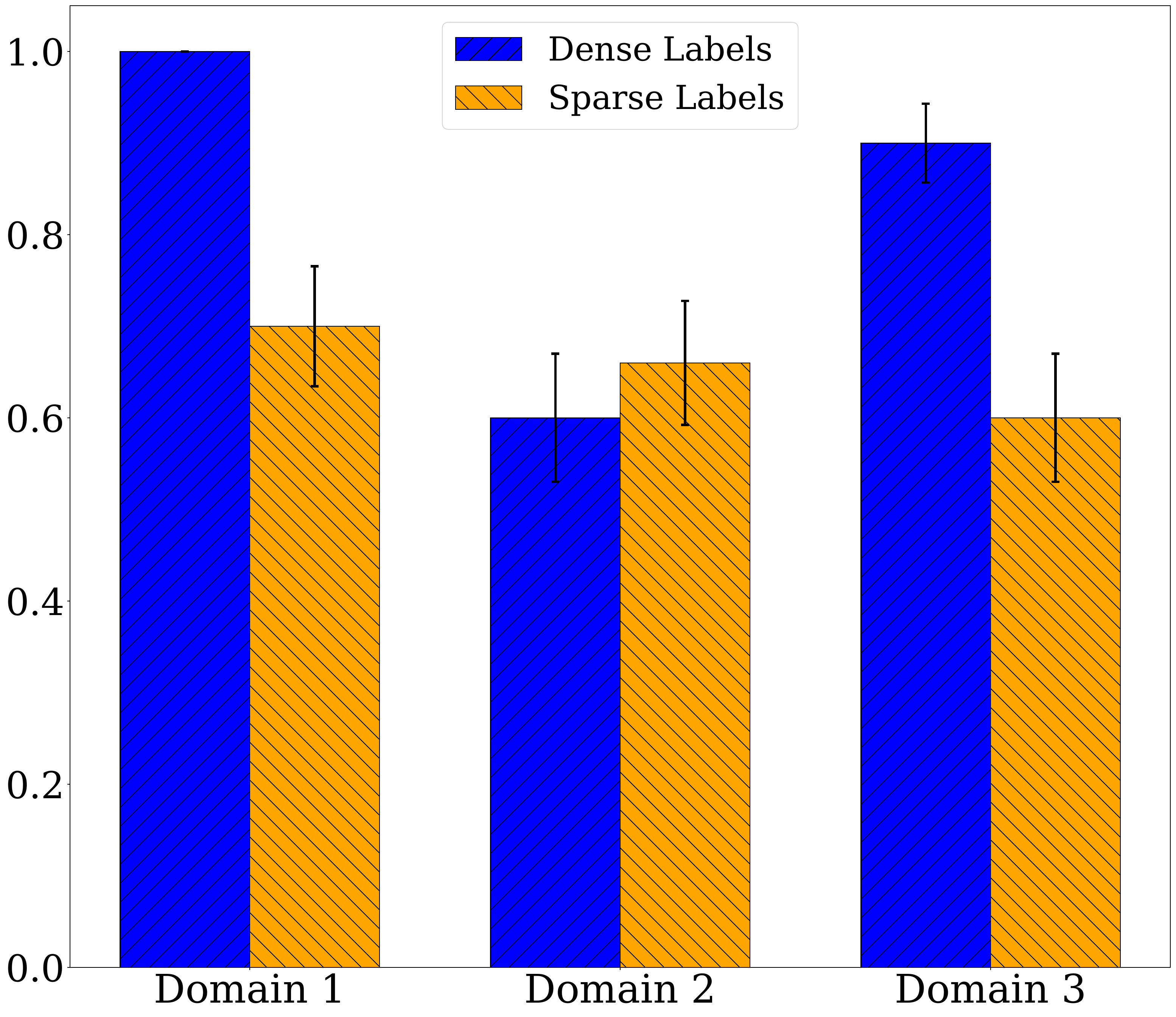}
        \caption{}
        \label{fig:lab_den}
    \end{subfigure}

    \caption{Comparison of Claude Sonnet 3.5 results across three experimental conditions: \textbf{(a)} Map Size: performance on original maps \textit{vs.} doubled maps over hard tasks; \textbf{(b)} Task Difficulty: performance on doubled maps over easy \textit{vs.} hard tasks; \textbf{(c)} Label Density: performance on original maps annotated with sparse \textit{vs.} dense labels over hard tasks.}
    \label{fig:claude_comparison_horizontal}
\end{figure}

The results indicate that accuracy decreases as map size increases in the overall domain analysis, with the VLMs performing better in smaller maps.
The difference is significant, as a T-test on trials with GPT-4o revealed a drop in accuracy ($t = 6.13$, $p < 0.0001$). This supports our hypothesis that accuracy decreases as map size increases. The results are reported in Figures~\ref{fig:gpt_comparison_horizontal} and~\ref{fig:claude_comparison_horizontal}. Both GPT and Claude exhibit a similar pattern of performance decline with larger maps.

\vspace{-8pt}

\newpage
When the map size is enlarged, it was observed that the VLM agents frequently had difficulties in figuring out door ownership. For instance, some of the generated plan included infeasible actions such as going through door D from room R, where the door (D) does not belong to room R. This is a strong indication that in large floor plans, it is challenging for the VLMs to reason about door ownership. By comparison, such mistakes were rarely seen in small floor plans. 

We looked into the navigation plans generated by GPT-4o for hard tasks in the three original maps reported in Figure~\ref{fig:gpt_comparison_horizontal}, where dense labels were provided. Each map contained five tasks, and each task was run ten times. For instance, Domain 3 included the five navigation tasks of (`GARAGE', `BUREAU'), (`LINGERIE', `GARAGE'), (`GARAGE', `CHAMBRE ENFANT 1'), (`CHAMBRE PARENTS', `GARAGE'), and (`GARAGE', `WC').  Among the total 150 trials, GPT-4o was successful in 135 trials by generating navigation plans that correctly connected the initial and goal rooms. Among those 135 successful trials, 132 were optimal. In the 10 trials of the task (`GARAGE', `WC'), the VLM failed in one trial, and reported suboptimal plan D4–D3–D2–D1 three times, while the optimal path D4–D11–D1 was reported six times. For all other successful trials, the reported plans were optimal.

\vspace{-4pt}
\subsubsection{Hypothesis 2 (task difficulty)}

\textls[-4]{The second hypothesis investigates how task difficulty impacts accuracy. Easy tasks involve straightforward navigation between rooms, while hard tasks, require traversing multiple intermediate rooms, making the tasks more complex. We compared the performance between doubled maps over easy tasks and hard tasks.}

The accuracy of the GPT-4o models generally decreased with increased task difficulty, as more complex tasks led to lower accuracy in two out of the three maps. For example, a T-test on \texttt{doubled\_map\_2} indicated a drop in accuracy with $t = 2.88$ and $p = 0.0047$. The results from both VLMs generally support our hypothesis that more complex tasks lead to lower accuracy, as navigating through multiple rooms introduces additional challenges. One example task is detailed in Table~\ref{tab:path_garage_wc}.

\begin{table}[H]
\centering
\caption{An example navigation plan generated by GPT-4o in Map 1 shown on the left of Figure~\ref{fig:all_maps}. 
The initial location is ``Terrrasse Couverte'' and the goal location is ``Chambre~1''. 
This navigation task requires a sequence of nine actions. 
GPT-4o achieved 0.96 success rate in this map on similar hard navigation tasks, which demonstrates great promises for VLM-based map parsing research. 
}
\begin{tabular}{p{4.8cm}p{7.8cm}}
\toprule
\textbf{Number} & \textbf{Action} \\
\midrule
1 & ApproachDoor(D8) \\
2 & OpenDoor(D8) \\
3 & GoThrough(D8) \\
4 & ApproachDoor(D7) \\
5 & OpenDoor(D7) \\
6 & GoThrough(D7) \\
7 & ApproachDoor(D4) \\
8 & OpenDoor(D4) \\
9 & GoThrough(D4) \\
\bottomrule
\end{tabular}
\label{tab:path_garage_wc}
\end{table}

\begin{figure}[t]
    \centering
    \includegraphics[width=0.85\linewidth]{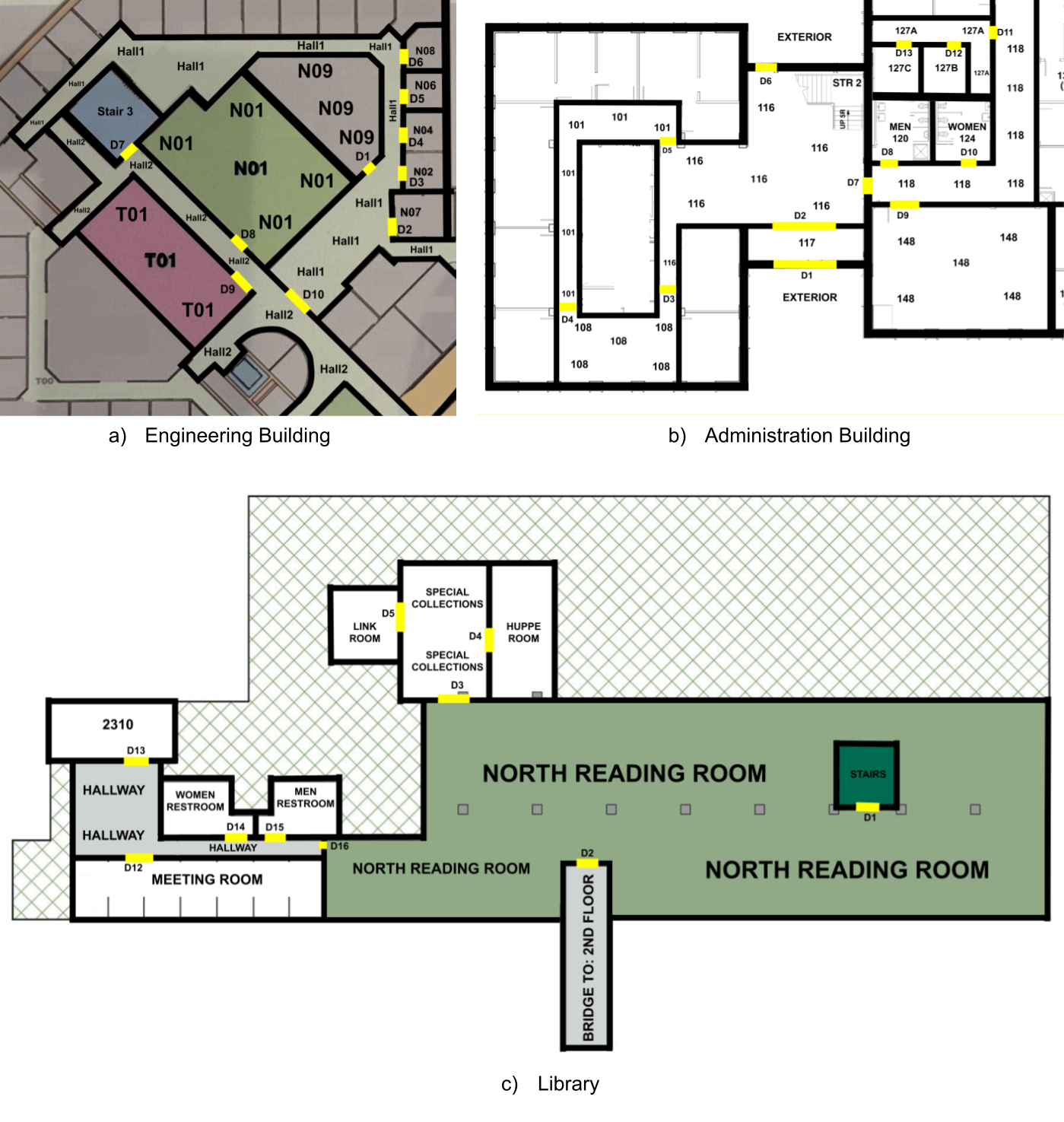}
    \vspace{-6pt}
    \caption{Three additional maps from the college campus. \textbf{(a)} Engineering Building; \textbf{(b)} Administration Building; \textbf{(c)} Library.}
    \label{fig:new-maps}
\end{figure}

Difficult tasks required the VLMs to generate navigation plans to go through three doors, which required strong reasoning capabilities about room connectivity and graph topology. One typical failure case was that VLMs insisted on directly entering a room between the start and goal rooms, when there is no direct access to the room in the middle. Consider the floor plan in Figure~\ref{fig:new-maps}a. \textls[+20]{When the robot started in room “N09” and aimed to go to room “Stair 3”, VLMs frequently suggested the robot to directly enter room N01. While N01 is immediately adjacent to both N09 and Stair 3, the robot cannot directly enter N01 from N09, because the only door of N01 connects to Hall2 instead of N09. Such observations indicate that VLMs had difficulty in reasoning about topological graphs of room connectivity through doors. }

\subsubsection{Hypothesis 3 (label density)}

The third hypothesis examines the impact of label density on accuracy. This is evaluated by comparing the performance of original maps from sparse-label and dense-label datasets over hard tasks. Dense labels provide more contextual information, while sparse labels offer minimal context, making the navigation task more challenging.

The accuracy of the GPT-4o models significantly improved with dense labels compared to sparse labels across all maps. For instance, a T-test on \texttt{original\_map\_1} showed a significant improvement with $t = 10.72$ and $p < 0.0001$. This result indicates that dense labels substantially enhance accuracy, supporting our hypothesis that dense labels provide crucial contextual information, enabling the models to navigate more effectively. The improvement in accuracy with dense labels was consistent across all maps for GPT-4o, highlighting the importance of label density in successful navigation.

Without dense labeling, VLMs frequently made mistakes in reasoning about door ownership and room connectivity through doors. For instance, door D16 in Figure~\ref{fig:new-maps}b connects two rooms “NORTH READING ROOM” and “HALLWAY”, where the former is very large. We found that the VLMs sometimes wrongly believed that D16 connects “MEN RESTROOM” and “HALLWAY” because those two room labels are closer to the door label. To facilitate VLMs to reason about such door ownerships, we added two additional labels for the reading room, and one of them was placed next to door D16, where dense labeling was found useful for map parsing and navigation task planning.

These findings suggest that our approach can effectively handle a range of navigation challenges while demonstrating the critical importance of label density, task difficulty, and map size in determining overall performance.

\vspace{-4pt}
\subsubsection {Raw \textit{vs.} processed floor plan maps}
We evaluated VLM performance on raw floor plan maps for the hard tasks. As shown in Figure~\ref{fig:raw-maps}, these raw images contain various irrelevant elements that are not essential to the spatial layout, such as furniture symbols, dimension lines, and scale bars. The results, presented in Table~\ref{tab:rawmaps}, show that accuracy is substantially lower when using raw maps compared to the processed sparse-label maps. This supports our claim in Section 3.1 
that raw maps contain distracting visual elements which may confuse the VLM in understanding the overall layout.

\begin{table}[t]
    \centering
    \caption{Average accuracy results comparing processed and raw floor plan maps across three domains for hard tasks.}
    \begin{tabular}{p{2.1cm}p{2.1cm}p{2.1cm}p{2cm}p{2cm}p{2.1cm}p{2cm}}
    \toprule
        \multirow{2}{*}{Map Type} & \multicolumn{3}{c}{GPT-4o} & \multicolumn{3}{c}{Claude Sonnet 3.5} \\
        \cmidrule(lr){2-4} \cmidrule(lr){5-7}
        & Domain 1 & Domain 2 & Domain 3 & Domain 1 & Domain 2 & Domain 3 \\
    \midrule
        Processed Maps & 0.24 & 0.84 & 0.22 & 0.68 & 0.66 & 0.34 \\
        Raw Maps & 0.03 & 0.21 & 0.05 & 0.08 & 0.18 & 0.22 \\
    \bottomrule
    \end{tabular}
    \label{tab:rawmaps}
\end{table}

\subsubsection{Diverse floor plans}
To evaluate the generalizability of our conclusions across different environments, we extended our experiments to three additional floor plans selected from the home institution of this article’s authoring team, including the Engineering Building, the Library, and the Administration Building. The Engineering Building is also the environment where we conducted the real-world demonstration, described in \mbox{Section 5} on Hardware Demonstration. These three floor plans are illustrated in Figure~\ref{fig:new-maps}. The results, presented in Table~\ref{tab:new-maps-results}, show that the VLMs performed well on the processed maps.

\begin{table}[H]
    \centering
    \caption{Average accuracy of VLM performance on three college campus floor plans.}
    \begin{tabular}{ccc ccc ccc}
    \toprule
        \multirow{2}{*}{Task Difficulty} & \multicolumn{3}{c}{GPT-4o} & \multicolumn{3}{c}{Claude Sonnet 3.5} \\
        \cmidrule(lr){2-4} \cmidrule(lr){5-7}
        & Engineering & Administration & Library & Engineering & Administration & Library \\
    \midrule
        Easy Task & 0.92 & 0.90 & 0.92 & 0.78 & 0.70 & 0.94 \\
        Hard Task & 0.80 & 0.94 & 0.90 & 0.54 & 0.62 & 0.90 \\
    \bottomrule
    \end{tabular}
    \label{tab:new-maps-results}
\end{table}

\section{Hardware demonstration}

\begin{figure}[H]
    \centering
     \includegraphics[width=0.95\textwidth]{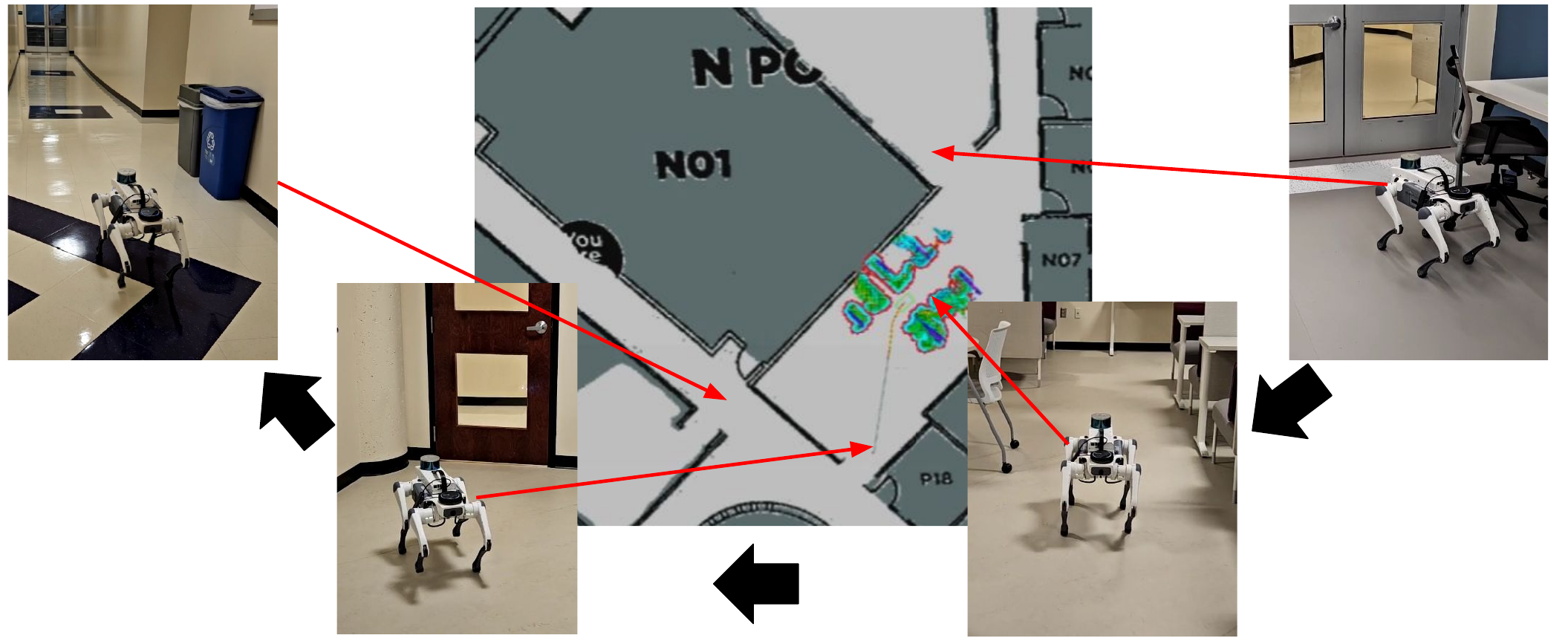}
    \caption{The robot localizes directly on the floor plan image and performs a navigation task. The occupancy grid map that the robot used for localization and navigation was also derived from the floor plan map captured by the robot. We manually removed the labels that would otherwise be interpreted by the robot as obstacles. From right to left, the robot was performing (1) Going through Door N09, (2) Approaching Door N00, (3) Opening Door N00, and (4) Approaching Door T01.}
    \label{fig:eb_photos}
\end{figure}

Our VLM-based planning and navigation system is demonstrated on a DEEP Robotics Lite3 quadruped robot. An image of the floor plan of a building on a college campus is captured from the robot's camera. This image is then edited as described in Section 3 
to make it suitable for the VLM query. Another version of the raw image with space whited out is directly used for robot localization, as seen in Figure~\ref{fig:eb_photos}. The VLM is then queried with the edited photo, and a navigation plan consisting of a sequence of navigation and door opening actions is generated. The robot executes this navigation plan, to move from the robotics lab (room N09), to a classroom (room T01). The robot localizes itself directly on a grayscale version of the floor plan, without the need for generating an accurate occupancy grid via SLAM. 

\vspace{-8pt}
The robot's software for localization, obstacle avoidance and navigation, including actions of approaching and going through doors, was constructed using an existing software codebase for building-wide robot intelligence~\cite{khandelwal2017bwibots}. 
The codebase supported labeling door areas on both sides of a door. 
Whenever the robot gets inside either of the door areas, it is believed that the robot successfully completes the action of approaching the door. 
The action of going through a door is realized by moving from one of the door areas to the other side of the same door. 
Our robot cannot open doors by itself and door opening actions are realized by repeatedly saying ``Please help me open the door” to seek human help until the two areas of the door become accessible to each other. 
We used the Lite3 robot's locomotion skills that rely on its default Model Predictive Controller (MPC). 
The robot used standard methods from the ROS community for navigation, including the dynamic window approach (DWA) for obstacle avoidance, A* for global path planning, and Adaptive Monte Carlo Localization (AMCL) for localization using grayscale occupancy-grid map images. 
The robot successfully avoided the obstacles, asked for doors to be opened as needed, and achieved the desired navigation goal, as illustrated in Figure~\ref{fig:eb_photos}.

\section{Conclusion and future work}

In this work, we introduce a novel task, named map parsing, that is unexplored in the VLM literature while pointing to the foundation of mobile robotics. 
We demonstrate remarkable performance of two VLMs on map parsing tasks, as applied to robot navigation. 
We develop a VLM-based planning system that generates navigation plans directly from a floor plan image and validated our approach through experiments on a floor plan dataset and on hardware, demonstrating the feasibility of VLM-driven navigation across different environments. 

While our results show that this approach is viable, several challenges remain, opening up avenues for further research. 
While the process of floor map enhancement was manually done in this article, one can easily automate the process using the following steps to reproduce our results. First, all labels and text outside the floor plans should be removed. One can use segmentation models~\cite{zhang2023faster, ravi2024sam, long2015fully} to locate those items for removal. 
After that, the formats of walls including thickness and line style should be unified. 
For instance, walls with windows are frequently indicated using hollow boxes, and it’s necessary to replace them with solid line segments. 
After that, all door icons should be unified. While we did not perform a comprehensive comparison, yellow rectangles were observed effective in our implementation. 
Since the navigation plans frequently include actions of going through doors~\cite{thomaz2023robots}, it is necessary to give each door a unique ID by adding a unique door name next to each door icon in the floor plans. 
Finally, dense labeling of rooms is necessary by copying and pasting room names to open and near-door areas of large rooms.

Another direction for automating the floor map enhancement pipeline is to leverage the state-of-the-art image editing tools. 
For instance, one can prompt ChatGPT~\cite{yang2023set} or Gemini~\cite{comanici2025gemini} with the raw map as image prompt and with editing instructions as the text prompt (e.g., replacing door symbols in this floor plan with solid yellow rectangles). 
To further improve the performance, one can employ chain-of-thought (CoT)~\cite{wei2022chain} style reasoning along with few-shot examples to prompt the model to follow the manual sequence of operations. 
Those operations include removing external labels and annotations, normalizing wall and window representations, standardizing door icons, and assigning unique IDs to each door. 
An alternative way of automating the map enhancement process is to leverage segmentation tools such as SAM 2~\cite{ravi2024sam} to mark out objects in the floor map and assign symbolic labels to the identified objects. 
Accordingly, walls can be thickened and door colors can be changed automatically, as a follow-up step. 
Automating the process of map enhancement can be an interesting and important future work direction to enhance this research from the engineering perspective. 

Finally, researchers can investigate strategies to improve the VLM's ability to handle larger and more complex maps, e.g., outdoor environments~\cite{pmlr-v205-shah23b}. 
Public parks and college campuses have been studied as the playground of mobile robot navigation~\cite{karnan2022socially,song2024vlm}, and they are usually provide visitors with maps near the entrances. 
Those maps have rich labels and spatial information. 
While we only evaluated our approach in indoor environments in this paper, we believe our approach can be furthered towards outdoor environments with larger areas and complex structures. 
VLMs and other transformer based autoregressive models have a host of known issues, such as hallucinations and biases~\cite{zhang2023sirenssongaiocean,10.1145/3442188.3445922,siWhyToxicMeasuring2022,garg-ramakrishnan-2020-bae} that can be addressed in robot planning, which deserves more attention too. 
For instance, VLMs might assume every home environment has a kitchen (which is wrong in places)~\cite{wu2019bayesian} and wrongly suggest a robot to go to a kitchen to get water in the kitchen that does not exist. 
To really bring our map parsing approach and the robot systems to end users, it is necessary to develop safeguards to ensure the output of VLMs do not introduce risks to people and the environments~\cite{ravichandran2025safety,robey2025jailbreaking}. 
This is particularly important when mobile robots are used for guiding people who are visually or physically challenged, e.g., robotic guide dogs~\cite{defazio2023seeing,kim2025understanding,hwang2024towards}, where this work was initially motivated by such application domains. 
We anticipate that this work will inspire further studies that expand upon the ideas on VLM-based map parsing in this paper. 

\section*{Data availability statement}
The data or datasets generated or analyzed in this study are available in GitHub, where the link is available on the project webpage: \url{https://sites.google.com/view/vlm-floorplan/}


\section*{Acknowledgments}
A portion of this work has taken
place at the Autonomous Intelligent Robotics (AIR) Group,
SUNY Binghamton. AIR research is supported in part by
the NSF (IIS-2428998, NRI-1925044), Ford Motor Company, 
DEEP Robotics, OPPO, Guiding Eyes for the Blind,
and SUNY RF.

\section*{Authors’ contribution}
DeFazio contributed to Methodology, Data curation, Formal analysis, Writing–review and editing, Software, Conceptualization, Investigation, Writing–original draft. 
Mehta contributed to Methodology, Data curation, Formal analysis, Writing–review \& editing, Software, Conceptualization, Investigation, Writing–original draft. 
Wang contributed to Methodology, Data curation, Formal analysis, Writing–review \& editing, Software, Conceptualization, Investigation, Writing–original draft. 
Yang contributed to Methodology, Data curation, Formal analysis, Writing–review \& editing, Funding acquisition, Supervision, Conceptualization, Writing–original draft. 
Blackburn contributed to Methodology, Data curation, Formal analysis, Writing–review \& editing, Funding acquisition, Supervision, Conceptualization, Writing–original draft. 
Zhang contributed to Methodology, Data curation, Formal analysis, Writing–review \& editing, Funding acquisition, Supervision, Conceptualization, Writing–original draft.


\section*{Conflicts of interests}
The authors declare no conflict of interest.

\bibliographystyle{aiasbst}
\bibliography{bibliography}

\pagebreak

\end{document}